\documentclass[journal]{IEEEtran}
\usepackage[T1]{fontenc}
\usepackage{textcomp}
\usepackage{tgtermes}   
\usepackage[numbers,sort&compress]{natbib}
\usepackage{multicol}
\usepackage{amsmath, amssymb}

\usepackage{amsthm}
\theoremstyle{definition}

\newtheorem{theorem}{Theorem}

\makeatletter
\AtBeginDocument{%
  \renewenvironment{proof}[1][\proofname]{\par\pushQED{\qed}\normalfont\topsep6\p@\@plus6\p@\relax\trivlist\item[\hskip\labelsep\bfseries#1\@addpunct{.}]\ignorespaces}{\popQED\endtrivlist\@endpefalse}%
}
\makeatother
\usepackage{graphicx}
\graphicspath{{Imgs/}{../Imgs/}{figures/}}
\usepackage{pifont}
\usepackage{booktabs}
\usepackage{threeparttable}
\usepackage{multirow}
\usepackage{array}
\usepackage{algorithm}
\usepackage{algpseudocode}
\algrenewcommand\algorithmicindent{1.2em}
\usepackage{xcolor}
\usepackage{placeins}
\usepackage{stfloats}      
\usepackage{afterpage}     
\usepackage[caption=false,font=footnotesize]{subfig}
\usepackage[bookmarks=true]{hyperref}

\newif\ifanon\anonfalse 
\begin{document}

\title{\LARGE\bf Deploying Embodied Model with Online Residual Imitation}

\ifanon
\author{Author Names Omitted for Anonymous Review. Paper-ID [add your ID here]}
\else
\author{Qi Guo$^{1}$, Entong Dong$^{1}$, Wenshuo Gou$^{1}$, Tong Yang$^{12}$, Jie Tang$^{2}$
\thanks{Corresponding author: Tong Yang (\textit{tong.yang@z.ai})}
\thanks{$^1$ Z-Lab, Z.AI}
\thanks{$^2$ Department of Computer Science and Technology, Tsinghua University}
}
\fi

\maketitle

\begin{abstract}
This paper proposes a new paradigm for online residual imitation learning that aims to reduce the burden of deployment-side adaptation, thereby enabling the deployment of larger-scale pretrained action models for which per-site retraining is impractical.
This contrasts with the conventional pipeline which separates training, deployment, and retraining into distinct stages. 
The main difficulty to achieving residual imitation learning has been recognised as the causal confusion problem: A residual module often degenerates into an independent behaviour clone that cancels the base model. 
It is shown in this paper that the solution is pretty simple: Do not feed the base model's action to the residual head. 
The advantages it takes are significant: no access to the base model's internals, no reward function design, and no separate retraining phase, with improvements taking effect within hours whilst the operator continues correcting.
This paradigm applies to all kinds of action models, including vision-language-action (VLA) models, world action models (WAMs), and non-fine-tunable black-box foundation models with embodied capabilities. 
Simulated experiments on selected LIBERO-Plus and BEHAVIOR-1K tasks show that all three classes of action models benefit from the proposed paradigm, with success rates rising by 19 to 36 percentage points, and GPT-6 Astra reaching 99.6\% on LIBERO-Plus with prompt engineering alone. 
A real-world socket-insertion task further validates the paradigm, raising the success rate of $\pi_{0.5}$ from 10.0\% to 53.3\% with two hours of human data, outperforming HG-DAgger and HIL-SERL. 
Moreover, the paradigm has been applied to public real-world embodied challenges, the LeHome folding challenge and the EBiM manipulation challenge, winning first place in both. 
\end{abstract}

\IEEEpeerreviewmaketitle

\section{INTRODUCTION}\label{sec:intro}

\begin{figure}[t]
\centering
\includegraphics[width=\columnwidth]{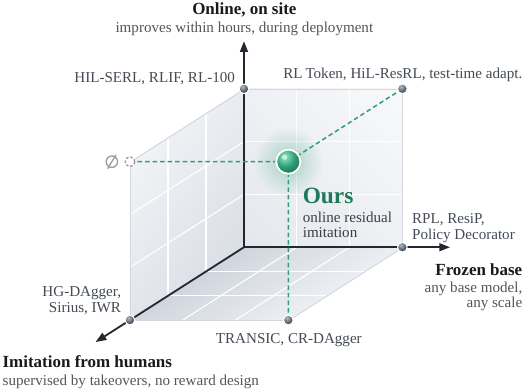}
\caption{
Positioning of the proposed paradigm. 
Each axis is one property that a deployment-side adaptation algorithm should satisfy: \textit{online, on site}, so that a correction takes effect within hours whilst the operator keeps correcting; \textit{imitation from humans}, so that no reward need be designed; and \textit{frozen base}, so that any base model at any scale is compatible. 
Existing methods lie on a single axis or on a plane spanned by two axes, satisfying at most two of the three properties. 
The corner that combines online updates with human supervision but lacks a pre-trained base is empty ($\varnothing$), because updating the full model online is too slow to take effect on site, and a small model trained from scratch cannot carry a language-conditioned manipulation task. 
The corner where all three hold has also remained empty, because feeding the base action to the residual module as input causes causal confusion. 
Removing it is what makes online residual imitation realisable, and this is the aim of this work. 
}
\label{fig:fig1}
\end{figure}

Ask a robot to assemble a part on a factory line, pack a box in a warehouse, or tidy a kitchen at home, and today's action models, built atop large language models~\cite{kim2024openvla}, can already carry out most of the tasks in an end-to-end fashion: reaching, grasping, and carrying. 
Yet the final step of physical contact with the world, when millimetres or degrees decide the outcome, is precisely where they fail to fulfil the tasks~\cite{kwon2026correctvla}: a connector may land just to one side of its socket, or a gripper may close a moment too soon and let the object slip free. 
Embodied tasks exhibit a wide spectrum of precision requirements, and even within a single task different subprocesses may demand different levels of manipulation precision. 
In a peg-in-hole task, for instance, only the aligning phase requires far higher precision than other time steps. 
However, this variation in precision demand is never taken into account during training. 
The embodied capabilities that a foundation model acquires during pre-training (prior to domain-specific fine-tuning) can be viewed as arising from maximum likelihood estimation over the empirical distribution of the pre-training dataset. 
Such an objective rewards capturing the broad motion pattern across generalised scenes, not being precisely accurate in any one of them. 
This opens a large gap in the amount of data required to move a general policy, within one specialised requirement, from producing the right directional trend of motion to actually meeting the precision requirement at the point where precision matters most. 
For a well-pretrained model, a handful of samples or even in-context learning suffices to produce the correct motion trend, whereas fine-tuning on far more data still does not guarantee that the output meets the precision requirement.
Moreover, spending large amounts of training data to chase the point of highest precision in a specialised task in fact causes a markable drop in generalisation at actual deployment.
This is a painful dilemma rooted in the training mechanism itself (minimising maximum-likelihood error on a given dataset), which cannot be systematically resolved except by strengthening the base model's capability.
Precision is the most conspicuous shortfall, but not the only one: lighting that differs from training, a slightly moved camera, or an object added to or removed from the scene can likewise make an otherwise competent policy fall short on site. 

On the other hand, there is a logical inversion between the offline training pipeline and such shortfalls, which only manifest, together with their error feedback, at deployment. 
They can only be improved through training, yet they tend to surface only on site. 
Training-side data and evaluation can rarely cover every scene actually encountered on site, so the shortfall becomes apparent only once the system is genuinely deployed. 
Every current framework therefore consists of a tight coupling between the training side and the deployment side: once a problem is discovered at deployment, data must be collected and the system taken back to the training side for retraining, whether by fine-tuning on freshly collected demonstrations~\cite{houlsby2019adapter}~\cite{hu2022lora}~\cite{wang2025vlaadapter} or by merging operator takeovers into the training set~\cite{ross2011dagger}~\cite{kelly2019hgdagger}~\cite{liu2022sirius}. 
This retrain-and-redeploy cycle in practice slows the development pace on both sides.
It is therefore believed that the only reasonable approach is to separate general-purpose generalisation ability from task-specific needs: Without touching the model's general capability, one waits until the error is actually observed on site at deployment and only then design a small, local correction module.
This practice learns a correction online and on site, on top of a policy that is already deployed and whose weights no longer change. We call this practice \textit{deployment-side adaptation}.

Compared with the residual learning paradigm built on reinforcement learning (RL)~\cite{yuan2024policydecorator}~\cite{rltoken}~\cite{hilresrl2026} which has so far dominated deployment-side correction, learning the residual by imitating the operator spares two costs at once: exploration on real hardware with the risk and the personnel it ties up, and a reward function that is task-specific and hardest to specify precisely where contact is fine. 
Imitation-based residual learning does exist~\cite{zhou2026corewam}~\cite{fang2026pfd}, but it is trained offline and therefore forgoes what online learning offers: no separate training-deployment cycle, and improvements that are perceivable immediately. 
Test-time and deployment-time adaptation~\cite{zhao2026retrievesteer}~\cite{lee2026salt}~\cite{lee2026selfcomp} are online, but their correction signal comes from the policy's own perception. 
Systematic errors, at fine contact or under a shifted scene, are precisely what the policy cannot perceive and what an on-site operator must point out instead. 
That said, one might have expected an online residual head supervised by the difference between the operator's action and the base's action to have closed this gap long ago. 
Prior work~\cite{ross2011dagger}~\cite{kelly2019hgdagger}~\cite{xu2025compliantresidual} has moved in this direction, yet one natural design choice, feeding the base action to the residual head as an input, carries a defect that explains why online residual imitation has not materialised. 

The main observation of this paper is intuitive. 
Suppose the residual head receives both the base action and the observations, and is supervised by the difference between the operator's action and the base action. 
The reason for the intervention is hidden in the observations, perhaps the operator sees that the peg is about to be inserted off-centre, whereas the base action is an explicit signal, so the head learns to correct the base action \textit{because of} the base action rather than because of that hidden reason. 
This is causal confusion in imitation learning~\cite{dehaan2019causal}~\cite{codevilla2019limitations}~\cite{wen2020copycat}. 
The paradigm proposed in this paper therefore excludes the base action from the input of the residual head, and this single choice is what allows the three properties in Fig.~\ref{fig:fig1} to hold at once. 
It learns from the operator's takeovers by imitation, so no reward needs designing. 
It learns online whilst the operator continues to correct, so the improvement is observable on site within hours. 
It leaves the base model untouched, so any base at any scale is compatible. 
This paradigm is not limited to adding an independent residual module: For a foundation model that can only be called through its interface, writing the takeover signal into its context brings the same kind of improvement.

The novel contributions of this paper are listed as follows: 

\begin{enumerate}
\item \textbf{The first solution to imitation-based online residual learning. }
Eliminating this causal confusion failure mode opens up a new avenue for imitation learning of the human-demonstration residual at the deployment site without degraded performance. 
\item \textbf{An integrated human-in-the-loop framework for deployment-side adaptation. }
The practical merit of this framework is its rapid and observable learning progress. 
Being applicable to both VLA and WAM, it allows human operators to monitor policy improvement and issue updated corrections when needed. 
An implementation that applies the proposed paradigm to non-fine-tunable LLMs is also provided. 
\item \textbf{Championship-level performance in the real world.} Extensive simulated and realistic benchmarking results demonstrate the state-of-the-art performance of the proposed algorithm. 
In particular, the end-to-end framework has been deployed in the EBiM competition~\cite{ebim2026} and the LeHome competition~\cite{li2026lehome}\footnote{This is achieved together with the team in China Merchants Lion Rock AI Lab. }, winning first place. 
\end{enumerate}

The remainder of this paper is organised as follows. 
Section~\ref{section:related_works} reviews existing literature.
Section~\ref{section:problem_formulation} formulates the problem definition.
Section~\ref{section:method} details the proposed deployment-side adaptation paradigm.
The online training of the whole framework is presented in Section~\ref{section:training}. 
The application of the residual updating idea on foundation models is discussed in Section~\ref{section:fm}. 
Extensive simulated and realistic experiments in Section~\ref{section:experiments} evaluates the practical efficiency of the proposed algorithm, with final conclusive findings gathered in Section~\ref{section:conclusion}. 

\section{RELATED WORKS}\label{section:related_works}

\subsection{Residual Correction}
Superimposing a learned correction on a fixed controller dates back to feedback error learning~\cite{miyamoto1988feedback} in 1988.
Given a policy model or a controller, residual policy learning~\cite{silver2018rpl}~\cite{johannink2019residual} adds a residual network on top of it. 
Among them,~\cite{silver2018rpl} zero-initialises the residual network so that the composite system coincides with the base at the start of training. 
Almost all residual models are trained by reinforcement learning (RL). 
RL Token~\cite{rltoken} incorporates $L_2$ loss to restrict the deviation of the actual output from the model's original output. 
Policy Decorator~\cite{yuan2024policydecorator} sets an upper bound on the magnitude of the residual output. 
ResiP~\cite{ankile2024resip} adopts closed-loop corrections to the action chunks of a diffusion policy~\cite{chi2023diffusion} executed open-loop. 
Residual off-policy RL~\cite{ankile2025residualoffpolicy} trains the residual with an off-policy critic. 
Self-Correcting VLA~\cite{selfcorrecting2026} constructs a dense reward based on predictions of future states and task progress. 
Closest to our setting is HiL-ResRL~\cite{hilresrl2026}, which trains a residual on the output actions of a VLA with human-in-the-loop RL. 
However, its reward signal is hard to design, and real-robot exploration remains costly to be realised in the real world. 

Another line of work adds no residual and updates the policy itself directly. 
Q-VGM~\cite{qvgm2026} turns critic gradients into a velocity-matching target for the action expert. 
RL-100~\cite{lei2025rl100} post-trains diffusion policies with real-robot reinforcement learning. 
SOP~\cite{pan2026sop} proposes grouped online co-learning of multiple robots. 
The RECAP strategy~\cite{pistar06recap} feeds deployment-time expert interventions, demonstrations and on-policy data together into advantage-conditioned RL over the whole VLA. 
In summary, training a residual module with reinforcement learning has proven a valid strategy, despite its heavy burden of reward design. 
This is in contrast to the paradigm to be proposed in this work, which is imitation-based and involves no reward design. 

\subsection{Human Intervention as a Learning Signal}\label{sec:related-hil}
Treating human takeovers as a learning signal is an established route~\cite{celemin2022iil}. 
Existing methods differ in how the takeover data are used. 
One class updates the whole policy. 
Methods following the idea of DAgger~\cite{ross2011dagger}, such as HG-DAgger~\cite{kelly2019hgdagger}, IWR~\cite{mandlekar2020iwr}, and Sirius~\cite{liu2022sirius}, add the takeovers to the dataset and re-train the policy offline, so an improvement does not take effect until the next round of training is complete.
RLIF~\cite{luo2024rlif} and HIL-SERL~\cite{luo2024hilserl} instead feed the takeovers into reinforcement learning, which however requires exploration on the robot.
Recent work in this class is concerned with the source and reliability of the interventions rather than with how they are used: 
UniIntervene~\cite{deng2026uniintervene} replaces most human interventions with an agent, Max-Q selective imitation~\cite{wang2026maxq} imitates interventions selectively and keeps improving on top of them, and the method of~\cite{zhao2025suboptimal} learns from sub-optimal interventions.
The other class keeps the base policy fixed and learns a residual from the corrections. 
TRANSIC~\cite{jiang2024transic} labels the residual with the difference between the robot's state before and after a whole intervention.
Compliant Residual DAgger (CR-DAgger)~\cite{xu2025compliantresidual} admits a compliant interface that receives the operator's corrections represented as external force and trains a residual module on this force. 
In FlowCorrect~\cite{welte2026flowcorrect}, the operator superimposes an offset on the base model's action rather than taking over, and the adapter is trained with the corrected action as the target. 
FlowDAgger~\cite{murray2026flowdagger} adapts generative policies in latent space from interventions. 
All of these methods update the residual in batches between deployments, after a round of corrections has been collected, which are essentially offline techniques. 
In contrast, the residual in this paper is updated continuously after each episode of takeover, whilst the operator keeps correcting. 

\subsection{Causal Confusion in Imitation Learning}\label{sec:related-causal}

The causal confusion problem has in fact been encountered in many prior works, though not always under this name. 
~\cite{spencer2021feedback} even denied it and attributed it instead to a covariate shift driven by feedback.
The best-known instances are the inertia problem~\cite{codevilla2019limitations}, where a driving policy given the vehicle speed learns that a stopped car stays stopped, and the copycat problem~\cite{wen2020copycat}, where a policy given the observation history learns to repeat its previous action. 
In robot manipulation, distractors in demonstrations mislead visual policies~\cite{ramirez2024recon}~\cite{dai2025civil}, generalist policies bind actions to camera viewpoints and ignore the instruction~\cite{xing2025shortcut}, and VLAs are largely insensitive to language perturbations~\cite{liberoplus2025}, which CofactVLA~\cite{zhang2026cofactvla} attributes to the dense visual stream overriding the sparse instruction. 
Each instance has so far needed its own fix~\cite{wen2021keyframe}~\cite{chuang2022copycat}~\cite{park2021oreo}, or the general but expensive remedy of querying the expert online as in DAgger~\cite{ross2011dagger}. 
None of these has been shown to transfer to another instance, and whether it is called causal confusion or covariate shift, the family remains a standing hazard of imitation learning rather than a solved problem. 
The instance that this paper deals with belongs to the same family, but its structure is different. 
In the classical instances, the confounder, be it the vehicle speed or the previous action, is an input that happens to predict the label.
Here, by contrast, the base action is not only an input to the residual head but also a term of the label, since the label is the difference between the operator's action and the base action. 
The head can therefore reproduce that term exactly by copying its input, and the shortcut is not a side effect of limited data but the exact solution of the imitation problem, which no amount of data removes.
This structure is what allows the instance to be stated and proven as a theorem in Section~\ref{sec:naive}. 

\providecommand{\abase}{a_{b}}
\providecommand{\ah}{a_{h}}
\providecommand{\abaset}{a_{b,t}}
\providecommand{\aht}{a_{h,t}}
\providecommand{\dbase}{\delta_{b}}
\providecommand{\dbaset}{\delta_{b,t}}
\providecommand{\dhu}{\delta_{h}}
\providecommand{\dhut}{\delta_{h,t}}

\section{PROBLEM STATEMENT}\label{section:problem_formulation}

Let $\pi$ be a robotic base model, defined as
\begin{equation}\label{eq:base}
\pi: (\{I_t^k\}_{k=1}^K, s_t, L) \mapsto a_{b, t},\quad s_t, a_{b, t} \in \mathbb{R}^d,
\end{equation}
where $I_t^k$ is the $k$th camera observation at time step $t$, $s_t$ is the robot's state, $L$ is the language instruction, and $a_{b, t}$ is the action output, with the subscript $b$ denoting that it is generated by the ``base'' model. 
Without loss of generality, we assume that both $s_t$ and $a_{b, t}$ consist of the joint angles together with the gripper opening. 
In this case, the formulation is the most succinct because the action values and the state values are directly subtractable. 
Practically speaking, should the base model produce other types of representations, such as SE(3) end-effector movements or delta end-effector movement, the joint targets are obtained through corresponding transformation techniques.
Policies with chunked action output are also valid~\cite{zhao2023act}~\cite{chi2023diffusion}, whilst the detailed formalism is omitted for clarity. 

\begin{figure}[t]
\centering
\includegraphics[width=0.75\columnwidth]{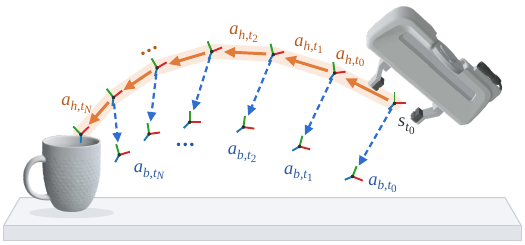}
\caption{Demonstration of the notations used in this paper. }\label{fig:notation}
\end{figure}

At a certain time step, the human operator may begin to take over control. 
Such interventions may occur multiple times within a single task execution episode, whilst only one of the continuous interventions is formulated here for clarity. 
As shown in Fig.~\ref{fig:notation}, denoting the starting time step of the intervention as $t_0$, the takeover spans the consecutive time steps $t_0$, $t_1$, $\cdots$, $t_N$, with $t_N$ being the last step that the human operator provides a takeover command. 
During this period, the action actually executed is that of the operator rather than of the base model. 
It is denoted as $a_{h, t_i}$, with the subscript $h$ standing for ``human'', and is expressed in the same joint space as $a_{b, t_i}$. 
The robot states under takeover are accordingly denoted as $s_{t_0}$, $s_{t_1}$, $\cdots$, $s_{t_N}$. 
Since the takeover is continuous, these time steps are simply
\begin{equation}
t_i = t_0 + i,\ i = 0, \cdots, N,
\end{equation}
and the notation $t_i$ is kept for brevity. 
At each of these steps, the base model still computes its intended action $a_{b, t_i}$, which is overridden by the operator's action. 

The \textit{online residual imitation learning} problem is to design a paradigm satisfying the following conditions
\begin{enumerate}
\item The base model $\pi$ is frozen. 
\item The overall action output is the composition of the base model's action and a residual correction, i.e., 
\begin{equation}
a_t = a_{b, t} + \Delta(\cdots), 
\end{equation}
where $\Delta$ denotes the residual module to be designed. 
\item The operator's takeover is the only external supervision. I.e., $\Delta$ is learned solely from the data collected during takeovers, 
\begin{equation}
\big\{\big(i,\ t_i,\ \{I_{t_i}^k\}_{k=1}^K,\ s_{t_i},\ a_{b, t_i},\ a_{h, t_i}\big)\big\}_{i=0}^{N},
\end{equation}
No reward and no autonomous exploration are involved. 
\item $\Delta$ is updated during deployment as takeovers accumulate, without an offline training cycle, so that takeovers from earlier episodes are reflected in later ones. 
\end{enumerate}
The problem is considered solved when the action output meets the operator's expectations, typically reflected in a reduction in the operator's takeovers and an improvement in task performance.

\section{PROPOSED DEPLOYMENT-SIDE ADAPTATION}\label{section:method}\label{sec:method}

\subsection{Common-sense Supervision Signal}\label{sec:naive}
The most intuitive imitation learning label is to set the complementary value $\Delta^*_{t_i}$ as the learning target, 
\begin{equation}\label{eq:label-naive}
\Delta^*_{t_i} = a_{h, t_i} - a_{b, t_i},\quad i \in \{0, \dots, N\},
\end{equation}
This choice, however, is problematic, which is formulated as the theorem below. 

\begin{figure}[t]
\centering
\includegraphics[width=\columnwidth]{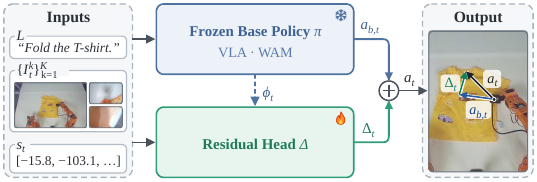}
\caption{Demonstration of the deployment of the proposed paradigm. }\label{fig:deployment}
\end{figure}

\begin{figure*}[t]
\centering
\includegraphics[width=\textwidth]{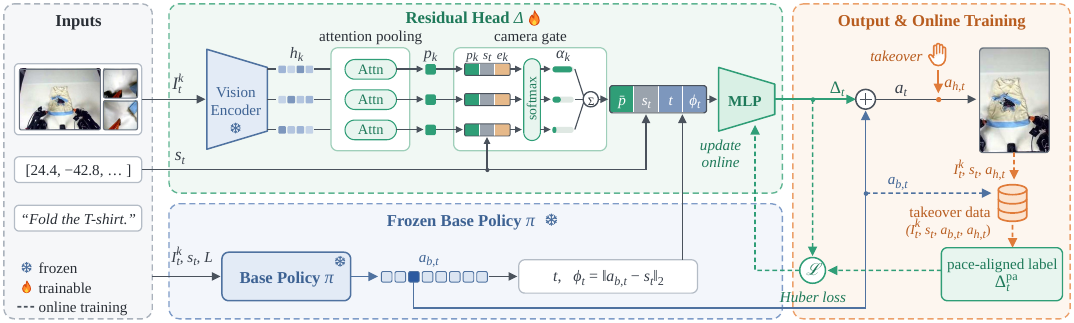}
\caption{Structure of the residual head and its online training. 
}
\label{fig:arch}
\end{figure*}

\begin{theorem}[Degeneration into Behavior Cloning]
\label{thm:confound}
Let the residual module $\Delta$ take the base action as one of its inputs, i.e., $a_{t_i} = a_{b, t_i} + \Delta(a_{b, t_i}, \cdots)$, where $\cdots$ denotes the remaining inputs $\{I_{t_i}^k\}_{k=1}^K$ and $s_{t_i}$, and let $\mathcal{F}$ be the family of all functions of $(a_{b, t_i}, \cdots)$, so that $\Delta\in \mathcal{F}$.
Then, for any training set and any per-sample loss $\mathrm{Loss}$, the residual problem
\begin{equation}
\min_{\Delta\in\mathcal{F}}\ \sum_i \mathrm{Loss}\big(\Delta(a_{b, t_i}, \cdots)-\Delta^*_{t_i}\big)
\label{eq:prob-res}
\end{equation}
and the behaviour-cloning problem, in which a policy $f\in\mathcal{F}$ maps the same inputs $(a_{b, t_i}, \cdots)$ directly to the action to be executed,
\begin{equation}
\min_{f\in\mathcal{F}}\ \sum_i \mathrm{Loss}\big(f(a_{b, t_i}, \cdots)-a_{h, t_i}\big),
\label{eq:prob-bc}
\end{equation}
have the same optimal value, and $\Delta$ solves Eqn.~\eqref{eq:prob-res} if and only if the policy $f=a_{b, t_i}+\Delta$ solves Eqn.~\eqref{eq:prob-bc}.
\end{theorem}
\begin{proof}
By Eqn.~\eqref{eq:label-naive}, for every $i$,
\begin{equation}
\Delta(a_{b, t_i}, \cdots)-\Delta^*_{t_i}=a_{b, t_i}+\Delta(a_{b, t_i}, \cdots)-a_{h, t_i},
\label{eq:identity}
\end{equation}
so the objective of Eqn.~\eqref{eq:prob-res} evaluated at $\Delta$ equals that of Eqn.~\eqref{eq:prob-bc} evaluated at $f=a_{b, t_i}+\Delta$.

Consider the mapping 
\begin{equation}
\Phi: \mathcal{F}\rightarrow \mathcal{F},\ \Delta\mapsto f=a_{b, t_i}+\Delta .
\end{equation}
$\Phi$ is well-defined, because $f$ is a function of $(a_{b, t_i}, \cdots)$ whenever $\Delta$ is.
It is also a bijection, since it has the inverse $f\mapsto f-a_{b, t_i}$, and $f-a_{b, t_i} \in \mathcal{F}$ because $a_{b, t_i}$ is among the inputs of $\Delta$.
Moreover, by Eqn.~\eqref{eq:identity}, $\Phi$ preserves the objective value. 
A bijection that preserves the objective value makes the two problems identical up to a relabelling of the candidates in $\mathcal{F}$. 
Therefore, their optimal values coincide, and a module $\Delta$ solves Eqn.~\eqref{eq:prob-res} if and only if $\Phi(\Delta)$ solves Eqn.~\eqref{eq:prob-bc}. 
\end{proof}

This theorem reveals that causal confusion is inevitable once the residual head receives the base action.
What the operator does, $a_{h, t_i}$, is caused by the scene (the object, the gripper, the task, etc.), and not by what the base model intended, since $a_{b, t_i}$ is overridden and never executed. 
The label $\Delta^*_{t_i}=a_{h, t_i}-a_{b, t_i}$ therefore consists of two parts of different nature: $a_{h, t_i}$, which depends on the scene, and $-a_{b, t_i}$, which is merely an input of the head with its sign flipped.
The second part costs the head nothing, since it can be copied from the input. 
What the head actually has to learn is the first part, namely the operator's full action from the images and the state.
The residual head is thus not learning a residual term to the base action but the action itself, with $a_{b, t_i}$ passed through and cancelled.
This is a vision-based action generation problem that a lightweight head trained on a handful of takeovers cannot solve--and it arises at exactly the states where the far larger base model has already failed. 
The base model's capability is discarded rather than built upon.

\subsection{New Paradigm}

Our solution to avoid causal confusion is therefore not to feed the base model's action into the residual head, only its norm to indicate the magnitude of the residual, meaning 
\begin{equation}\label{eq:head-io}
\Delta_t=\Delta\big(\{I_t^k\}_{k=1}^K,\ s_t,\ \phi_t, t\big).
\end{equation}
where 
\begin{equation}\label{eq:steplen}
\phi_t = \big\| a_{b, t} - s_t\big\|_2
\end{equation}
as shown in Fig.~\ref{fig:deployment}. 
The residual head receives images and proprioceptive state, and outputs a correction that is added to the base action. 
This suffices to avoid causal confusion and to solve the online residual learning problem, and is the only modification we make to the deployment of this new paradigm.
Details for the structure of this residual head and its training are provided in the next section. 

\section{ONLINE ADAPTATION}\label{section:training}

\subsection{Residual Head Structure}
The structure of the residual head and its online training are provided in Fig.~\ref{fig:arch}. 
The image inputs are encoded by the frozen visual encoder DINOv3~\cite{dinov3} and then compressed into a single vector $p_k$ by attention pooling~\cite{jaegle2021perceiver}:
\begin{align}
h_k&=\mathrm{DINOv3}(I_t^k),\ k = 1, \cdots, K\label{eq:encode}\\
p_k&=\mathrm{AttnPool}(h_k),\ k = 1, \cdots, K\label{eq:pool}
\end{align}
The camera gate scores each camera with a linear layer, normalises the scores with softmax, and sums the features to obtain a fused visual feature $\bar{p}$:
\begin{align}
z_k &= \mathrm{Linear}\big([\,p_k;\;s_t;\;e_k\,]\big),\ k = 1, \cdots, K\label{eq:gate}\\
\alpha_k &= \frac{\exp z_k}{\sum_{j=1}^{K}\exp z_j},\ k = 1, \cdots, K\label{eq:gate-weight}\\
\bar{p} &= \sum_{k=1}^{K} \alpha_k\,p_k,\label{eq:gate-feat}
\end{align}
where $e_k$ is a learnable embedding corresponding to the $k$th camera. 
As such, different images are assigned different importance, and the weighting is fully automated. 
Finally, the residual value is generated by a two-layer MLP, 
\begin{equation}
\Delta_t=\mathrm{MLP}\big([\,\bar{p};\;s_t;\;t;\;\phi_t\,]\big).
\label{eq:head}
\end{equation}
The authors believe that this is one of the simplest implementations to the residual head. 

\subsection{Training Label Construction}\label{sec:anchor}

Without causal confusion, directly applying $\Delta^*_{t_i}$ in Eqn.~\eqref{eq:label-naive} as the training label is feasible.  
However, it is observed that this label inherits the speed at which the operator moves the robot. 
When the operator makes fine adjustments during a takeover, especially near the target, the robot's step-wise displacement is often smaller than the step commanded by the base model. 
The operator's pace is a matter of teleoperation rather than of the task, and what is worth learning from a takeover is the positions through which the operator takes the robot, not how fast it moves. 
We therefore preserve the base model's step length when constructing the label and take only the path from the operator. 

\begin{figure}[t]
\centering
\includegraphics[width=0.8\columnwidth]{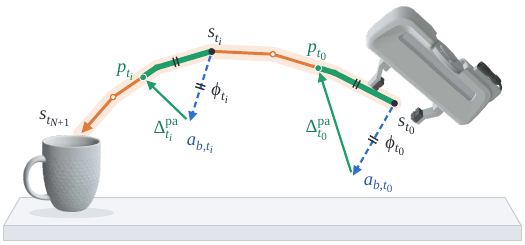}
\caption{Demonstration of the pace-aligned label. 
}
\label{fig:pace}
\end{figure}

Recall that the robot states during a continuous takeover are $s_{t_0},\,s_{t_1},\,\dots,\,s_{t_N}$, where $s_{t_N}$ is the state at which the operator issues the last takeover command. 
For simpler mathematical formulation, the position reached by this last command is denoted as
\begin{equation}
s_{t_{N+1}}\triangleq a_{h,t_N},
\label{eq:endpoint}
\end{equation}
so that the path travelled under the takeover is
\begin{equation}
s_{t_0},\,s_{t_1},\,\dots,\,s_{t_N},\,s_{t_{N+1}} .
\label{eq:path}
\end{equation}
When the takeover ends, the \textit{pace-aligned pose} $p_{t_i}$ at each step $t_i$ is computed as the point on this path whose travelling distance from $s_{t_i}$ equals the base model's step length $\phi_{t_i}$ in Eqn.~\eqref{eq:steplen}, as depicted in Fig.~\ref{fig:pace}. 
Denote the travelling distance from $s_{t_i}$ to $s_{t_m}$ along the path by
\begin{equation}
\Lambda_{i,m}=\left\{
\begin{aligned}
&0,\ & m = i, \\
&\sum_{j=i}^{m-1}\big\|s_{t_{j+1}}-s_{t_j}\big\|_2,\ &m\in\{i+1,\dots,N+1\},
\end{aligned}
\right.
\label{eq:pathlen}
\end{equation}
For the index $m$ such that
\begin{equation}
\Lambda_{i,m}\le \phi_{t_i}<\Lambda_{i,m+1},\qquad m\in\{i,\dots,N\},
\label{eq:segment}
\end{equation}
the $p_{t_i}$ is computed as
\begin{equation}
p_{t_i}=(1-\beta)\,s_{t_m}+\beta\,s_{t_{m+1}},\qquad \beta=\frac{\phi_{t_i}-\Lambda_{i,m}}{\big\|s_{t_{m+1}}-s_{t_m}\big\|_2}.
\label{eq:pose-label}
\end{equation}
If $\phi_{t_i}\ge\Lambda_{i,N+1}$, i.e., if the remaining path is shorter than the base model's step, no such $m$ exists and $p_{t_i}$ is set to the endpoint, $p_{t_i}=s_{t_{N+1}}=a_{h,t_N}$.

Finally, the training label $\Delta^{\mathrm{pa}}_{t_i}$ at time step $t_i$ is the difference between the pace-aligned pose and the base action,
\begin{equation}
\Delta^{\mathrm{pa}}_{t_i}=p_{t_i}-a_{b,t_i},
\label{eq:pace}
\end{equation}
called the \textit{pace-aligned label}. 
The opening and closing of the end-effector is not interpolated along the path; the end-effector dimension of every $p_{t_i}$ within the takeover segment takes the corresponding component of the operator's last command $a_{h,t_N}$.

\begin{figure*}[t]
\centering
\includegraphics[width=\textwidth]{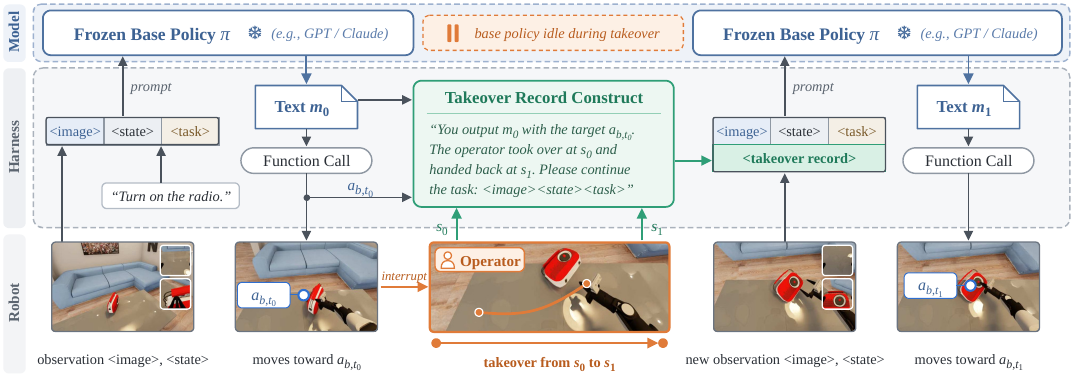}
\caption{
Pictorial demonstration of the human takeover for a generative foundation model. 
}
\label{fig:fm}
\end{figure*}

\subsection{Online Training}\label{sec:train}

During deployment, training samples are collected continuously. 
If a takeover is in progress, the training labels are accumulated. 
If no human intervention occurs, the action is deemed acceptable, and the label is therefore the model's own output $\Delta_t$, so that the residual head does not forget. 
Denoting this label as $y_t$, we have
\begin{equation}
y_t=\left\{
\begin{aligned}
&\Delta^{\mathrm{pa}}_{t_i},\ & \text{takeover},\\
&\Delta_t,\ & \text{otherwise}.
\end{aligned}
\right.
\label{eq:target}
\end{equation}

The residual head regresses the labels with a weighted Huber loss~\cite{huber1964}, regularised by an $L_2$ penalty on the correction itself:
\begin{equation}
\mathcal{L}=\sum_t\Big(w_t\,\mathrm{Huber}\big(\Delta_t-y_t\big)+\lambda\,\|\Delta_t\|_2^2\Big).
\label{eq:loss}
\end{equation}
The regulariser is necessary because otherwise training on the head's own outputs would allow the correction to drift arbitrarily at states without takeover data. 
It pulls the corrections at these states towards zero, so that the correction takes effect only near states where takeovers occurred.
The weight $w_t$ is determined by the stability of the operator's final command, 
\begin{equation}
w_t=\left\{
\begin{aligned}
&\max\Big(0,\;1-\frac{\|a_{h,t_N}-s_{t_N}\|_2}{\frac{1}{N+1}\sum_{i=0}^{N}\|a_{b,t_i}-s_{t_i}\|_2}\Big),\ &\text{takeover},\\
&w_{\mathrm{a}},\ &\text{otherwise},
\end{aligned}
\right.
\label{eq:weight}
\end{equation}
where $\|a_{h,t_N}-s_{t_N}\|_2$ is the norm of the operator's last command, the denominator is the average step length of the base model within the segment, and $w_{\mathrm{a}}$ is a constant (set to $0.3$ in our case). 
In other words, the weight of a takeover process is penalised if the robot ends with a moving state.  

Takeover samples and autonomous samples are stored separately, and each batch draws half from each. 
Within a batch, one-fourth of the training samples are drawn from the latest takeover episode, so that new takeovers take effect immediately. 
The remaining three-fourths are drawn uniformly from all existing samples. 
To avoid jumps in the correction between adjacent time steps, an exponential moving average is applied to $\Delta_t$ before it is added to the base model's action.

\section{RESIDUAL UPDATING WITH FOUNDATION MODELS}\label{section:fm}

Unlike the VLA and WAM models of the previous sections, embodied foundation models such as GPT-6 Astra~\cite{zhang2026astra} are not designed to output high-frequency continuous actions.
Their output is often not numerical but textual or consists of function calls, which a harness, i.e., a program that wraps the model and executes its tool calls, translates into executable motion. 
A typical output is a desired goal pose of the gripper, which is sparse in time, and the model is too slow for the operator to wait for its next output. 
Such models can often only be called through an interface and cannot be fine-tuned. 
The residual head of Section~\ref{section:training} therefore no longer applies: 
There are too few samples for it to converge during deployment, and there are no step-wise actions for it to correct. 
The proposed paradigm still applies, with the residual still learned from human takeovers, but its implementation differs.
The process is shown in Fig.~\ref{fig:fm}.

\begin{figure*}[t]
\centering
\includegraphics[width=\textwidth]{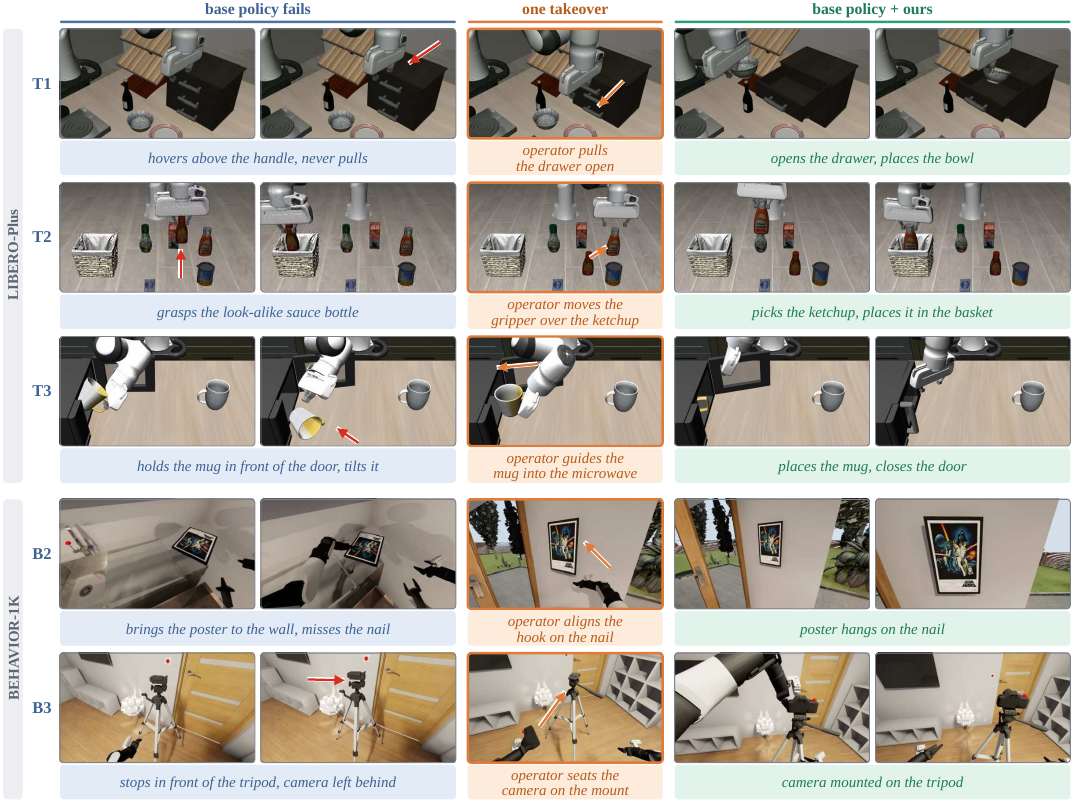}
\caption{Five typical failures in the simulated benchmarks and their corresponding human takeover with the corrected robotic behaviors. 
}
\label{fig:sim-stories}
\end{figure*}

Let $m_t$ be the text that the foundation model outputs at time step $t$. 
The harness parses from it the joint action target $a_{b, t}$ and drives the robot there through function calls. 
When the operator takes over, the harness does not record the whole takeover segment but only the state $s_0$ at which the takeover starts and the state $s_1$ at which it ends. 
These two states and the model's last output are written into the next prompt together with the new observations and the task instruction, as follows:
\begin{quote}\small\itshape
``You output $m_t$ with the target $a_{b,t}$. The operator took over at $s_0$ and handed back at $s_1$. 
Please continue the task: \textup{\texttt{<image><state><task>}}''
\end{quote}
The symbols are replaced with their values, and the three placeholders with the current images, the proprioceptive state and the task instruction. 
The correction thus enters the model through its input rather than being added to its output, and the model draws on it in the next inference. 
Although each takeover yields only two states, it is experimentally revealed that the foundation model learns from them quickly: Often, after two or three takeovers at the same failure, failures of that type largely no longer occur. 
Fine manipulation that requires continuous high-frequency correction from foundation models, rather than a few targets, remains beyond what this correction can offer.

\section{EXPERIMENTS}\label{section:experiments}
The proposed paradigm is evaluated in both simulated and real-world experiments. 
The simulated benchmarking in Section~\ref{sec:sim} is conducted on two benchmarks: LIBERO-Plus and Behavior-1K 2026.
On the most challenging sub-tasks of these benchmarks, where the base model ($\pi_{0.5}$) performs the worst, the proposed paradigm improves all baseline models (VLA, WAM and foundation model). 
In particular, the performance of GPT-6 Astra is improved to 99.6\% on LIBERO-Plus and 78\% on Behavior-1K 2026. 
Real-world socket-insertion experiments are presented in Section~\ref{sec:real}, where the proposed paradigm raises the success rate of $\pi_{0.5}$ from 10\% to 53.3\%. 
The proposed paradigm has also been applied to two open real-robot competitions, the ICRA LeHome Challenge 2026 and the EBiM Challenge 2026, winning first place in both. 
These are introduced in Section~\ref{sec:online}, and detailed ablation studies conducted in the EBiM scenario are provided in Section~\ref{sec:abl}.

\subsection{Simulated Comparisons}\label{sec:sim}

\begin{figure*}[!t]
\centering
\includegraphics[width=\textwidth]{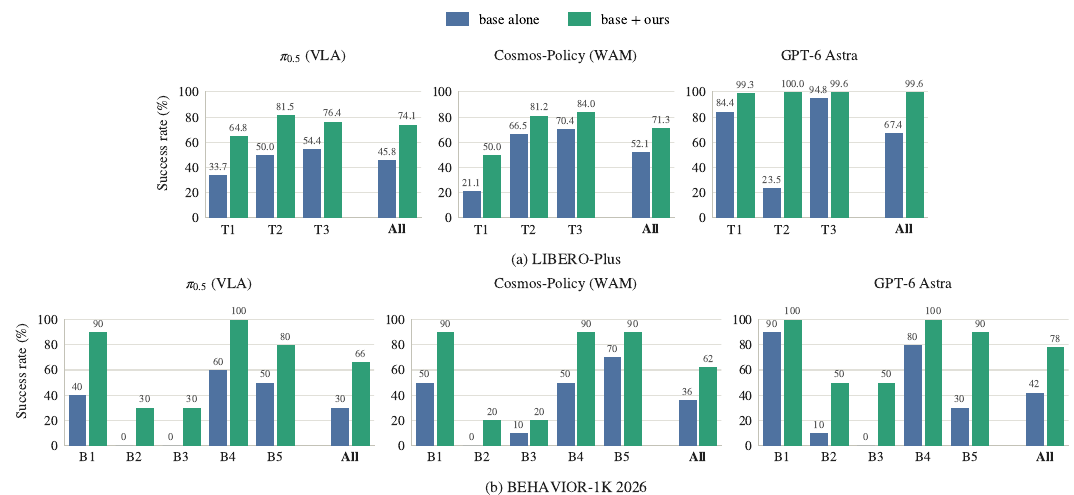}
\caption{Experimental results on simulated benchmarks LIBERO-Plus and Behavior-1K. 
}
\label{fig:main}
\end{figure*}

The scenes for simulated comparison are selected from the LIBERO-Plus~\cite{liberoplus2025} and BEHAVIOR-1K~\cite{behavior1k2026} benchmarks. 
To identify the most challenge sub-tasks, we trained $\pi_{0.5}$ on them and selected the perturbed variant-subtask pairs with the lowest success rates. 
This yields three challenging LIBERO-Plus sub-tasks under camera viewpoint variation and robot initial state variation: put the bowl into the top drawer (represented by T1), put the ketchup into the basket (T2), and put the mug into the microwave (T3), consisting of $260$ variants in total. 
This is split 7:3 into $182$ training variants and $78$ test variants.
Similarly, five challenging BEHAVIOR-1K tasks are selected, focusing on the final contact phase where existing base models are prone to failure: turning on a radio (B1), hanging pictures (B2), attaching a camera to a tripod (B3), installing a modem (B4), and composting waste (B5). 
Among the 20 public instances for each task provided by the Behavior-1K organisers, we train online residual imitation on the first 10 instances and evaluate on the last 10.
Human takeovers are collected only on the training set, and success rates are reported only on the test set.

Three representative action models are chosen in our experiments: $\pi_{0.5}$~\cite{pi05} for VLA, Cosmos-Policy~\cite{cosmos2025} for WAM, and GPT-6 Astra~\cite{zhang2026astra} for the generative foundation model. 
Human takeovers are implemented by a Pico VR controller in LIBERO-Plus and by an additional keyboard for mobile base control in BEHAVIOR-1K.
Case studies of how a failure is corrected after takeover are shown in Fig.~\ref{fig:sim-stories}.
Experimental results show that in all selected tasks the success rates of the three classes of base models improve significantly under the proposed paradigm, as recorded in Fig.~\ref{fig:main}. 
Most notably, $\pi_{0.5}$ on B2 and B3, Cosmos-Policy on B2, and GPT-6 Astra on B3 achieve success rates of $30\%$, $20\%$ and $50\%$ respectively, where they previously scored $0\%$. 
The proposed paradigm also improves the success rate of GPT-6 Astra on T3 from $94.8\%$ to $99.6\%$, which the authors believe is the lightest-weight approach (requiring only prompt engineering) for improving the performance of foundation models on embodied tasks.

\subsection{Real-world Experiments}\label{sec:real}

\begin{figure*}[t]
\centering
\subfloat[platform setting]{\includegraphics[width=0.285\textwidth]{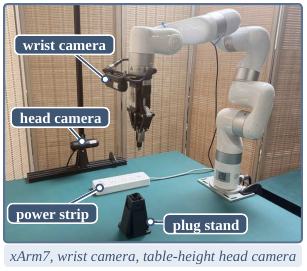}\label{fig:plug-a}}\hfill
\subfloat[task setting]{\includegraphics[width=0.322\textwidth]{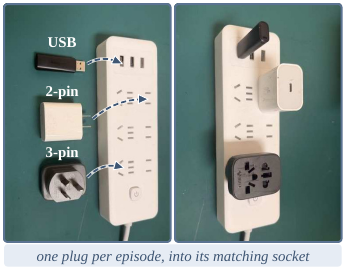}\label{fig:plug-b}}\hfill
\subfloat[human takeover]{\includegraphics[width=0.373\textwidth]{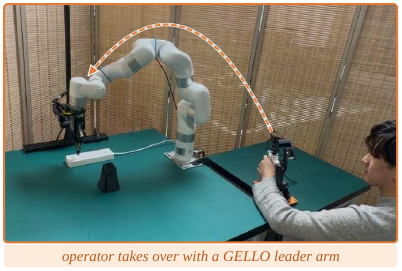}\label{fig:plug-c}}
\caption{Demonstration of the socket-insertion task. (a) The scene. (b) The three tasks: USB, two-pin and three-pin. (c) The GELLO teleoperation for takeover. }
\label{fig:plug}
\end{figure*}

Real-world validation is conducted on a socket-insertion task performed by a single-arm xArm. 
Three types of plugs are used--USB, two-pin and three-pin--and the task is to insert each plug into the corresponding socket of a power strip, as depicted in Fig.~\ref{fig:plug}.
Performance statistics are collected in Table~\ref{tab:plug}. 
The $\pi_{0.5}$ baseline is trained on $5$ hours of teleoperated demonstrations. 
The proposed paradigm is applied to the pretrained $\pi_{0.5}$ with 2 hours of human takeover effort. 
HG-DAgger~\cite{kelly2019hgdagger} merges the same 2 hours of human efforts into the training set and retrains the $\pi_{0.5}$ base. 
HIL-SERL~\cite{luo2024hilserl} is trained on 2 hours teleoperated data as demonstrations and then explores online on the real robot with a sparse success reward. 
With the same $2$ hours of human work, the proposed paradigm improves the average success rate from $10.0\%$ to $53.3\%$, correcting the frozen base online, which is more efficient than $\pi_{0.5}$ with HG-DAgger and more accurate than it (20.0\%). 
In this experiment, HIL-SERL fails to converge to a usable policy with the 2-hour data and achieves only a $3.3\%$ success rate.

\begin{table}[!t]
\centering
\caption{Comparison with human-in-the-loop learning methods on the real-robot socket-insertion task. }
\label{tab:plug}
\setlength{\tabcolsep}{3pt}
\begin{threeparttable}
\begin{tabular*}{\columnwidth}{@{\extracolsep{\fill}}l|c|ccc|c@{}}
\toprule
 & & \multicolumn{4}{c}{\textbf{Success (\%)} $\uparrow$} \\
\textbf{Method} & \textbf{Human h} $\downarrow$ & \textbf{USB} & \textbf{Two-pin} & \textbf{Three-pin} & \textbf{Ave} \\
\midrule
$\pi_{0.5}$ & 5 & 0.0 & 10.0 & 20.0 & 10.0 \\
\midrule
$\pi_{0.5}$+HG-DAgger & 2 & 10.0 & 10.0 & 40.0 & 20.0 \\
HIL-SERL & 2 & 0.0 & 10.0 & 0.0 & 3.3 \\
\textbf{$\pi_{0.5}$+ours} & 2 & \textbf{40.0} & \textbf{50.0} & \textbf{70.0} & \textbf{53.3} \\
\bottomrule
\end{tabular*}
\begin{tablenotes}[flushleft]
\item[] ``Human h" denotes the amount of time that a human operator takes.  
\item[] Each statistic data is the success rate over $10$ trials.
\end{tablenotes}
\end{threeparttable}
\end{table}

\subsection{Deployment in Robot Competitions}\label{sec:online}

\begin{figure*}[t]
\centering
\includegraphics[width=\textwidth]{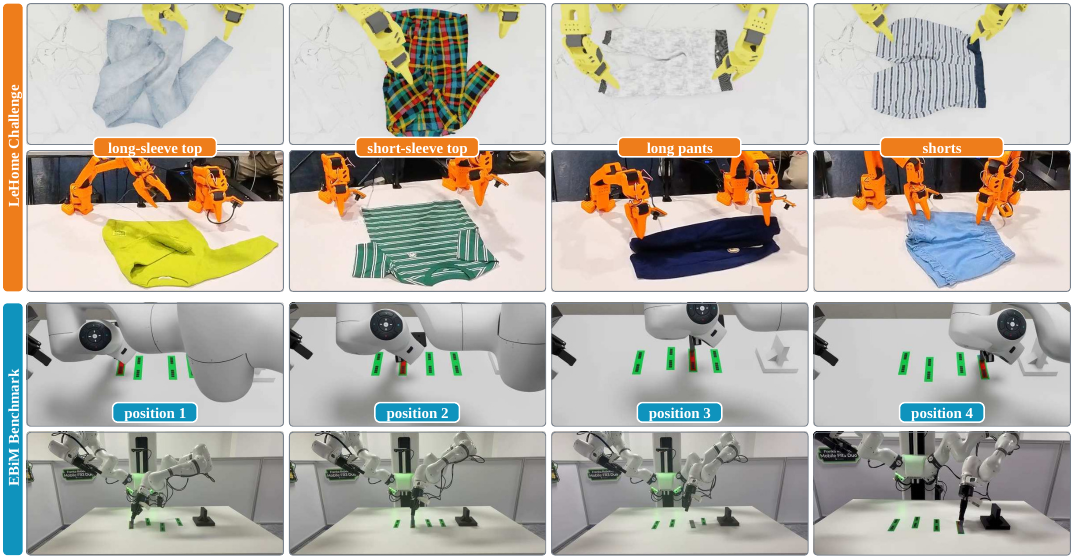}
\caption{Video stills of the LeHome and EBiM competitions. Upper rows: the simulated and real-world garments that were folded in LeHome. 
Lower rows: the simulated and real-world thermal pad placement to four positions in EBiM. } 
    \label{fig:competitions}
\end{figure*}

The proposed paradigm is also applied in two real-robot competitions. 
Illustrations of these competitions are shown in Fig.~\ref{fig:competitions}.
In the ICRA LeHome Challenge~\cite{li2026lehome}, an action model (we adopt $\pi_{0.5}$ again) is pretrained on four types of garments, long-sleeve tops, short-sleeve tops, long trousers and shorts, using SO-101 manipulators with three cameras. 
The score is given by the flatness and alignment of the folded garment, out of a maximum of $960$ points. 
We obtained $895$ points. 
In another real-world robotic challenge, EBiM~\cite{ebim2026}, we applied the proposed paradigm to Task 2, which uses a Mobile FR3 Duo robot (not yet publicly sold) to attach a deformable thermal pad to a target area. 
The IoU of the placement is the metric of this competition, as shown in Fig.~\ref{fig:abl-cases}(a). 
Apart from prior simulation-based training, all work regarding the real-world phase of the competition was conducted on-site on the organisers' robot. 
Ten trials were conducted at each of the possible target positions. 
With the proposed paradigm, the mean IoU of pad placement rose from 43.7\% to 75.1\%.

\subsection{Ablations}\label{sec:abl}

\begin{table}[t]
\centering
\caption{Ablation Studies}
\label{tab:ablation}
\begin{threeparttable}
\begin{tabular*}{\columnwidth}{@{\extracolsep{\fill}}l|cc@{}}
\toprule
\textbf{Setting} & \textbf{Success Rate (\%)} $\uparrow$ & \textbf{Coverage IoU (\%)} $\uparrow$ \\
\midrule
$\pi_{0.5}$ & 57.5 & 39.3 \\
\midrule
\textbf{$\pi_{0.5}$+ours} & \textbf{95.0} & \textbf{80.4} \\
with causal confusion & 40.0 & 50.9 \\
no pace-aligned label & 77.5 & 47.3 \\
no camera gate & 90.0 & 64.8 \\
\bottomrule
\end{tabular*}
\begin{tablenotes}[flushleft]
\item[] Each statistic data is the average over $10$ trials. 
\end{tablenotes}
\end{threeparttable}
\end{table}

\begin{figure}[t]
\centering
\subfloat[coverage (IoU)]{\includegraphics[width=0.510\columnwidth]{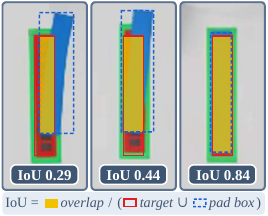}\label{fig:abl-cov}}\hfill
\subfloat[failure: no placement]{\includegraphics[width=0.476\columnwidth]{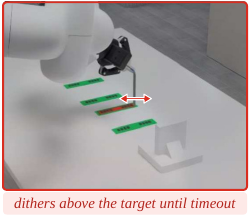}\label{fig:abl-fail}}
\caption{Demonstration of the coverage IoU and a failure case in the ablations.
}
\label{fig:abl-cases}
\end{figure}

Ablation studies are conducted on the EBiM Task 2 scenario. 
Because the competition schedule was very tight, we were unable to conduct ablations on the real robot and therefore performed them in simulation. 
In contrast to the vanilla pretrained $\pi_{0.5}$ and the proposed paradigm built upon it, three variants are compared: re-introducing causal confusion (by feeding $\pi_{0.5}$ action into the residual head), skipping the construction of pace-aligned labels (Eqn.~\eqref{eq:pose-label}), and removing the camera gate (Eqn.~\eqref{eq:gate}--Eqn.~\eqref{eq:gate-feat}).
Table~\ref{tab:ablation} reports the corresponding statistics.
The success rate metric counts the ratio of episodes in which the thermal pad overlaps the correct target region within the time limit, and the coverage IoU metric is the same as in the competition. 
A visual comparison of high and low coverage is shown in Fig.~\ref{fig:abl-cases}(a). 
Results show that the pretrained $\pi_{0.5}$ achieves a success rate of $57.5\%$ and a coverage IoU of $39.3\%$, which the proposed paradigm improves to $95.0\%$ and $80.4\%$, respectively.

With causal confusion, performance is even worse than that of the vanilla pretrained $\pi_{0.5}$, with a success rate of 40\%, because, as analysed in Section~\ref{section:method}, causal confusion causes the residual head to learn the whole action policy rather than the intended residual. 
In a large portion of the failed episodes, the gripper hovers above the target area making small back-and-forth adjustments until timeout. 
Such failures are shown in Fig.~\ref{fig:abl-cases}(b). 
With causal confusion eliminated, even when the human operator's label is sent directly for training, performance still improves to a success rate of $77.5\%$ and a coverage IoU of $47.3\%$.
Finally, designing camera gates is observed to indeed let the residual head focus on the most informative camera view, without which there is a small drop to a success rate of $90.0\%$ and a coverage IoU of $64.8\%$.

\section{CONCLUSION}\label{section:conclusion}
This paper has introduced a new online residual imitation learning paradigm for deploying pretrained action models in robotics, including VLA and WAM models and foundation models. 
Theoretical analysis has proven that the key lies in eliminating causal confusion, which is implemented simply by not feeding the base model's action into the residual head.
This gives birth to the proposed paradigm. 
Without separate retraining, reward design, or fine-tuning of the base action model, the authors believe the paradigm is sufficiently lightweight and general to suit a wide range of base action models and robotic tasks. 
Experimentally, the proposed paradigm demonstrates improved performance on two simulation benchmarks and on a real-world socket-insertion task, and won first place in two international robot competitions. 

\bibliographystyle{IEEEtran}
\bibliography{online_residual_imitation}

\end{document}